\documentclass{article}

    \usepackage[final,nonatbib]{neurips_2020}

\usepackage[style=numeric,sorting=none,doi=false,isbn=false,url=false,eprint=false,backend=biber]{biblatex}
\usepackage[utf8]{inputenc} 
\usepackage[T1]{fontenc}    
\usepackage{hyperref}       
\usepackage{url}            
\usepackage{booktabs}       
\usepackage{amsfonts}       
\usepackage{nicefrac}       
\usepackage{graphicx}
\usepackage{microtype}      
\usepackage{chngcntr}

\graphicspath{ {Images/} }

\title{Learning Canonical Transformations}

\author{%
  Zack ~Dulberg \\
  Princeton Neuroscience Institute\\
  Princeton, NJ 08544 \\
  \texttt{zdulberg@princeton.edu} \\
   \And
   Jonathan Cohen \\
   Princeton Neuroscience Institute \\
   Princeton, NJ 08544 \\
   \texttt{jdc@princeton.edu} \\
}

\begin{document}

\maketitle

\begin{abstract}
  Humans understand a set of canonical geometric transformations (such as translation and rotation) that support generalization by being untethered to any specific object. We explore inductive biases that help a neural network model learn these transformations in pixel space in a way that can generalize out-of-domain. Specifically, we find that high training set diversity is sufficient for the extrapolation of translation to unseen shapes and scales, and that an iterative training scheme achieves significant extrapolation of rotation in time. 
\end{abstract}

\section{Introduction}

Humans have a unique ability to generalize knowledge outside the scope of prior experience \cite{Chollet2019,lake2017building,Marcus2001}, while artificial agents struggle to apply knowledge to distributions outside the convex hull of their training data \cite{santoro2018measuring,lake2018generalization}. To achieve such generalization, humans seem to learn a set of primitive abstract structures, like the 1D ordinal scale \cite{summerfield2020structure} and the grid-like representation \cite{Hafting2005}. These structures can also be thought of as symmetry functions; transformations that are in some way invariant to the identity of their arguments. As a concrete example, we can imagine moving any object around in space, regardless of its identity. A fundamental question is how do humans learn symmetry functions?

We hypothesize that during development, infants learn a set of canonical transformations - the translation and rotation of objects - that are grounded in the sensorimotor system \cite{Barsalou2008}, and learned as a consequence of predicting the sensory results of primitive actions \cite{Battaglia2013}. The abilities to translate and rotate arbitrary objects then become our first abstract affordances \cite{Gibson2014}. Indeed, infants that spend more time playing with blocks are better at abstract mental rotation tasks \cite{Schwarzer2013}, and there is even evidence that motor action is crucial for the development of abstract social reasoning \cite{Sommerville2005,Woodward2009}. 

Our aim is to model part of this process by presenting a fully convolutional neural network \cite{lecun1995convolutional} with a 2-dimensional shape, and training it to predict the effect of a translation or a rotation of that shape in pixel space. This is analogous to predicting tactile or visual signals resulting from a simple movement or saccade \cite{rao1999predictive,wolpert1995internal}. We do not explicitly model motor actions, but rather transform the image by hand, and feed the result back to the model. We then test the extent to which such predictions generalize out of domain along dimensions such as shape, size, and time.  Evidence of out of domain (o.o.d.) generalization would suggest that the network has learned a symmetry function. 

We assume that, in order to learn symmetry functions, we must introduce principled inductive biases. The first is convolution itself, which has proven capable of learning representations corresponding to those found in the human brain \cite{Yamins2014}. Second, to constrain the network to learn a primitive function that can apply to any shape, we assume it requires exposure to a sufficiently diverse set of examples. We therefore operationalize and vary 'diversity' as the number of distinct shapes present in the training set. We then consider the effects of iteration during training, based on the idea that sequential applications of the same transformation should maintain the identity of an object (i.e. object permanence \cite{Piaget2006}). We find that diversity and iteration can trade off with each other to produce o.o.d. generalization of learned transformations.

\section{Method}

\paragraph{Environment}

All training stimuli were shapes contained within a 64x64 pixel grid space. We constructed irregular N-sided polygonal shapes by first sampling N angular distances between 0 and $2\pi$, and then sampling a radial distance (from a centroid with horizontal and vertical offsets $x$ and $y$) at each of these angles uniformly between 0 and a scale parameter $r$. This produced a set of vertices; pixels within the convexity of the vertices were set to 1, and pixels outside were set to 0. This produced a combinatorically large set of possible shapes. This procedure for shape generation is most similar to Attneave forms \cite{attneave1956quantitative} and also bears relation to the method of Fourier descriptors \cite{Zhang2005}, but was selected due its computational speed and interpretable manipulations of shape parameters. Once generated, each shape was used as the input to a fully convolutional neural network, and was transformed in one of the following ways to generate the target for training: for translation, shift 2 pixels to the right; for rotation, rotate $\frac{\pi}{25}$ radians clockwise. These transformations were hard-coded but meant to represent a set of innate primitive actions.

\paragraph{Model Architecture}

The model was a fully convolutional autoencoder. The encoder consisted of 3 convolutional layers, the first with 16 3x3 kernels, (stride=2, padding=1), the second with 32 3x3 kernels (stride=2, padding=1), and the last with 64 7x7 kernels (stride=1, padding=1). The decoder had 3 layers (padding=1, output padding=1) that inverted these operations with transposed convolution layers that were mirror images of the encoder layers (i.e. layers in the decoder had stride ½ when the corresponding layer in the encoder had stride of 2). This produced an output with the same dimensions as the input image. All layers were followed by rectified linear (ReLu) activations. Since the network was fully convolutional, it could also accept any input grid size. All weights were initialized using Xavier uniform initialization. 

\paragraph{Training}

The networks were trained by providing a randomly generated shape as input, and back-propagating the mean squared error (MSE) loss between the output of the network and the appropriately transformed target shape. We trained separate networks on translation and rotation. For translation, inputs were shapes with parameters (in pixel units) sampled from r $\in$ [5,7], x $\in$ [20,25] and y $\in$ [20,40], and targets were input shapes with x increased by 2. For rotation, inputs were shapes with parameters r $\in$ [7,10], x = 32 and y = 32, and targets were input shapes rotated clockwise by $\frac{\pi}{25}$ radians (compared to translation, these shapes were centered and slightly larger). Weight updates were performed with the ADAM optimizer, using a learning rate 5e-4, weight decay of 1e-5, and minibatch size of 32. Networks were all trained for 100,000 steps.  

In order vary diversity, we trained a separate set of networks for each of 100, 500, 1000, 10000 and 'inf' items in the training set ('inf' involved generating new images on the fly every batch, to approximate an infinite diversity of shapes). To vary iteration, we introduced a training variant that treated the network as an iterated function, based on the principle that $f(n_{t+1}) = f(f(n_t))$, where $f$ is a translation or rotation function and $n_t$ is a shape after $t$ transformations.  For each input, the final output was generated by $k$ applications of the network. In other words, the output was passed back through the network as input $k$ times, and the final output was compared with the target image (initial image transformed $k$ times, for example, $\frac{k\pi}{25}$  clockwise rotations). The accumulated gradient from these passes was then used to perform a single weight update. For each training batch, the integer $k$ was sampled uniformly between 1 and $M$, with a separate network trained for $M$ ranging from 1 and 9 to assess the effect of varying the amount iteration. While this procedure resembles recurrent neural networks like the LSTM \cite{Hochreiter1997}, it differs in that it accepts only a single input (whereas recurrent networks usually accept sequences) and it only propagates the output, rather than a hidden network state, so it is memory-less. This procedure has in fact been successfully used to generate rotations of 3D shapes \cite{Galama2019}. 

\paragraph{Testing}

At test time, the networks were presented with new shapes, and then repeatedly applied (outputs passed back as inputs) to assess the correspondence between each successive transformation and the correct shape at that time point (not to be confused with the iterative training procedure). Accuracy was defined as the sum of pixel errors expressed as a fraction of the area of the target shape. We used a selection of test datasets to assess out of domain extrapolation along dimensions like shape, size, and time (time defined as repeated transformation of a stimulus). Each test data set contained 500 shapes. 

Translation networks were tested far out of domain from the training set, using a set of shapes with a much larger overall radius parameter $r$ (50 pixels), with hollow-centers (to change the shape distribution), and presented in a 512x512 grid (a much larger scale). Rotation networks were tested over the same distribution as their training sets in terms of shape and scale, as they failed to extrapolate to very different shapes or a larger grid size. In the case of rotation, we focused on extrapolation in time; since networks were trained only on one, or a small number (1-9) of iterations, repeatedly transforming stimuli much longer than that at test time (up to 50 times) would reflect extrapolation. As well, due to the combinatorial size of the shape space, I.I.D. generalization rather than memorization was still necessary to succeed at this task. 

\section{Results}

The best translation network (Div. inf, It. 1) was capable of nearly perfect out of domain generalization (Fig. 1, Left), as it was able to translate novel, large shapes at previously unseen scales and locations. Shapes also maintained their identity with repeated transformations of this network (i.e. while many networks produced a reasonable output at Time = 1, the best networks were accurate even at Time = 50). Interestingly, all that was necessary for this capacity was the 'inf' diversity condition. In fact, with high diversity, iterative training slightly hurt performance (Fig. 1, bottom left). Conversely, there was an apparent trade-off at lower levels of diversity, where iteration could improve performance (As a concrete illustration of this, after 50 time steps, the Div. 500, It. 9 network achieved a mean accuracy of 0.9, while the Div. 10000, It. 1 network achieved slightly lower accuracy of 0.84).

\begin{figure}[h]
  \centering
    \resizebox{\textwidth}{!}{ }
  \includegraphics{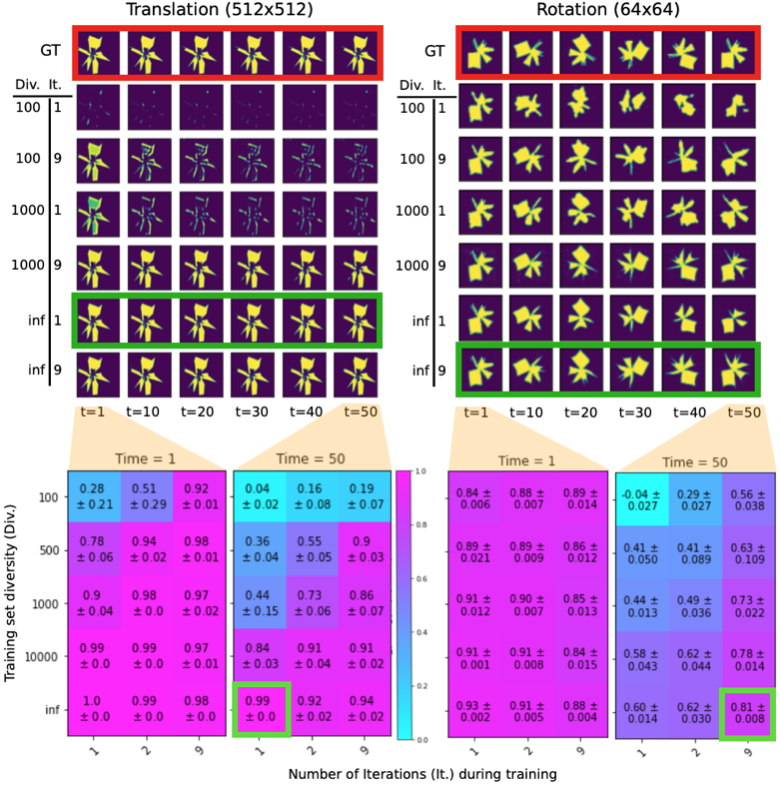}
  \caption{Left: Translation networks tested on 512x512 grid. Right: Rotation networks tested on 64x64 grid. Top: Repeated applications of selected networks after input of a single frame. Networks vary in diversity (Div.) and iterations (It.) during training. The red box shows the ground truth (GT) sequence. The green box indicates the best performing network. Bottom: Mean accuracy and standard error (of 3 identically trained networks) for a larger set of diversity and iteration values, displayed at the 1st and 50th time points. Accuracy ranges between 0 and 1 and is color coded. Green boxes refer to the same best performing networks as in the top panels.}
\end{figure}

The best rotation network (Div. inf, It. 9) was capable of extrapolation in time, as can be seen visually by comparing its outputs to ground truth images (Fig. 1, top right). The trade-off between iteration and diversity was also clear here as indicated by the diagonal colour gradient in the Time = 50 grid (Fig. 1, bottom right). However, unlike for translation, optimal performance at rotation was by maximizing both diversity and iteration during training. It is worth noting that greater iteration had the opposite effect at Time = 1 (it hurt performance). This indicates that iteration does not simply tighten the error tolerance of each network output to achieve greater stability in time (shown in Fig. A.1 for additional values of iteration). These iterative networks were also more stable in time than networks trained to rotate large angles in a single pass (Fig. A.2), so iterative networks rotate at smaller increments while maintaining greater accuracy over time.

\section{Discussion}

In this work, we tested the ability of auto-encoders to extrapolate learned transformations in pixel space. Using convolution and high enough training set diversity, translation networks learned the correct symmetry function, that is, the arbitrary shift of 1-valued pixels. This network was capable of significant extrapolation with respect to shape, scale, and time. Recent work has proven that simple ReLu MLP networks will extrapolate linear functions with enough diversity in their training set \cite{xu2020neural}, and here we provide empirical support outside the range of this theoretical result for convolutional networks in image space.  This result is also consistent with the diversity effect seen in human psychology, \cite{Osherson1990}, and with recent evidence that diversity is required for generalization in reinforcement learning tasks \cite{hill2019emergent}.

The rotation network did not extrapolate with regard to shape or scale, but a combination of iteration and diversity resulted in stable rotation far past the time horizon seen during training, or in other words, extrapolation in time. Prior work has investigated this kind of temporal extrapolation in recurrent networks, but solutions usually required baking in a conservation law of some sort \cite{cranmer2020lagrangian, greydanus2019hamiltonian}. Here, we propose that conservation of shape emerges from the symmetry implied by training the networks as iterated functions; as Noether’s theorem states, for any symmetric action, there is a corresponding conservation law \cite{Noether2005}. Closest to this work is the iterative generative adversarial network which learned 3D rotation using a similar iterative training technique; however, that work did not assess how different amounts of iteration during training impacted extrapolation in time \cite{Galama2019}. Iteration is also a plausible mechanism in humans - psychological data for mental rotation suggests that we transform objects iteratively (since larger angles of rotation elicit longer reaction times) \cite{Shepard1971}, and there is evidence that discrete temporal context updating by recurrent thalamocortical loops serves predictive learning in the brain \cite{o2014learning}. 

Finally, we found an interesting interaction between diversity and iteration, where each could partially make up for less of the other to produce o.o.d. generalization capacities. Based on all these results, we suggest that humans may use these strategies synergistically during development in order to learn canonical transformations. Intuitively, a child may optimally learn to predict the sensory effects of these transformations by both transforming many different objects as well as repeatedly transforming the same object. In this way, they are able to abstract canonical transformations away from individual object instances, while maintaining a sense of object permanence.

For future work, we aim to investigate why networks easily learned the correct translation symmetry function, which only required high training set diversity, but had difficulties learning a rotation function that could extrapolate along dimensions like shape or scale. We hypothesize this was because the translational invariance built into convolution was well aligned with the task of translation, but less so with rotation. We predict that using a rotationally-invariant convolution \cite{Cohen2016} might solve this problem, and are currently extending our analysis in that direction.

\section*{Broader Impact}

In machine learning, failures to extrapolate knowledge into unfamiliar domains severely limits the effectiveness of automation in areas like diagnostics or medical imaging \cite{Choudhary2020}, and results in unacceptable bias when, for example, a network that is trained on a majority population performs poorly on and under-represented one \cite{mehrabi2019survey}. At a time when society is experiencing an explosion in big data/automation in the context of systemic inequity, understanding the development of  extrapolative capacity is important given our goal to engineer fair and equitable machine learning models.

\section*{Acknowledgements}

I would like to thank Taylor Webb, Steven Frankland, Simon Segert, Randall O'Reilly, and Alexander Petrov for their helpful discussions and feedback.

\printbibliography 

@article{Noether2005,
archivePrefix = {arXiv},
arxivId = {physics/0503066},
author = {Noether, Emmy},
doi = {10.1080/00411457108231446},
eprint = {0503066},
issn = {15322424},
journal = {Transport Theory and Statistical Physics},
month = {mar},
number = {3},
pages = {186--207},
primaryClass = {physics},
title = {{Invariant Variation Problems}},
url = {http://arxiv.org/abs/physics/0503066 http://dx.doi.org/10.1080/00411457108231446},
volume = {1},
year = {1971}
}

@inproceedings{Cohen2016,
archivePrefix = {arXiv},
arxivId = {1602.07576},
author = {Cohen, Taco S. and Welling, Max},
booktitle = {33rd International Conference on Machine Learning, ICML 2016},
eprint = {1602.07576},
isbn = {9781510829008},
month = {feb},
pages = {4375--4386},
publisher = {International Machine Learning Society (IMLS)},
title = {{Group equivariant convolutional networks}},
url = {http://arxiv.org/abs/1602.07576},
volume = {6},
year = {2016}
}

@article{Barsalou2008,
author = {Barsalou, Lawrence W.},
doi = {10.1146/annurev.psych.59.103006.093639},
isbn = {9780824302597},
issn = {00664308},
journal = {Annual Review of Psychology},
month = {jan},
number = {1},
pages = {617--645},
publisher = {Annual Reviews},
title = {{Grounded cognition}},
url = {http://www.annualreviews.org/doi/10.1146/annurev.psych.59.103006.093639},
volume = {59},
year = {2008}
}

@article{Zhang2005,
author = {Zhang, Dengsheng and Lu, Guojun},
doi = {10.1016/j.imavis.2004.09.001},
issn = {02628856},
journal = {Image and Vision Computing},
month = {jan},
number = {1},
pages = {33--49},
publisher = {Elsevier Ltd},
title = {{Study and evaluation of different Fourier methods for image retrieval}},
url = {https://research.monash.edu/en/publications/study-and-evaluation-of-different-fourier-methods-for-image-retri},
volume = {23},
year = {2005}
}

@article{Schwarzer2013,
author = {Schwarzer, Gudrun and Freitag, Claudia and Schum, Nina},
doi = {10.3389/fpsyg.2013.00097},
issn = {16641078},
journal = {Frontiers in Psychology},
number = {},
publisher = {Frontiers Research Foundation},
title = {{How crawling and manual object exploration are related to the mental rotation abilities of 9-month-old infants}},
url = {/pmc/articles/PMC3586719/?report=abstract https://www.ncbi.nlm.nih.gov/pmc/articles/PMC3586719/},
volume = {4},
year = {2013}
}

@article{Shepard1971,
author = {Shepard, Roger N. and Metzler, Jacqueline},
doi = {10.1126/science.171.3972.701},
issn = {00368075},
journal = {Science},
number = {3972},
pages = {701--703},
pmid = {5540314},
title = {{Mental rotation of three-dimensional objects}},
volume = {171},
year = {1971}
}

@article{Sommerville2005,
author = {Sommerville, Jessica A. and Woodward, Amanda L. and Needham, Amy},
doi = {10.1016/j.cognition.2004.07.004},
issn = {00100277},
journal = {Cognition},
month = {may},
number = {1},
pages = {B1--B11},
pmid = {15833301},
publisher = {Elsevier},
title = {{Action experience alters 3-month-old infants' perception of others' actions}},
volume = {96},
year = {2005}
}

@article{Battaglia2013,
author = {Battaglia, Peter W. and Hamrick, Jessica B. and Tenenbaum, Joshua B.},
doi = {10.1073/pnas.1306572110},
issn = {00278424},
journal = {Proceedings of the National Academy of Sciences of the United States of America},
month = {nov},
number = {45},
pages = {18327--18332},
pmid = {24145417},
publisher = {National Academy of Sciences},
title = {{Simulation as an engine of physical scene understanding}},
url = {https://www.pnas.org/content/110/45/18327 https://www.pnas.org/content/110/45/18327.abstract},
volume = {110},
year = {2013}
}

@incollection{Gibson2014,
author = {Gibson, James J.},
booktitle = {The People, Place, and Space Reader},
doi = {10.4324/9781315816852},
isbn = {9781315816852},
pages = {56--60},
title = {{The theory of affordances (1979)}},
url = {https://books.google.com/books?hl=en&lr=&id=b9WWAwAAQBAJ&oi=fnd&pg=PA56&dq=gibson+theory+of+affordances&ots=KV3wzKoqzd&sig=p66pWxlCM3XzeryBbQmrWi8x7o0},
year = {2014}
}

@article{Woodward2009,
author = {Woodward, Amanda L},
doi = {10.1111/j.1467-8721.2009.01605.x},
issn = {0963-7214},
journal = {Current directions in psychological science},
month = {feb},
number = {1},
pages = {53--57},
pmid = {23645974},
publisher = {SAGE PublicationsSage CA: Los Angeles, CA},
title = {{Infants' grasp of others' intentions.}},
url = {http://www.ncbi.nlm.nih.gov/pubmed/23645974 http://www.pubmedcentral.nih.gov/articlerender.fcgi?artid=PMC3640581},
volume = {18},
year = {2009}
}

@article{Hafting2005,
author = {Hafting, Torkel and Fyhn, Marianne and Molden, Sturla and Moser, May Britt and Moser, Edvard I.},
doi = {10.1038/nature03721},
issn = {00280836},
journal = {Nature},
month = {aug},
number = {7052},
pages = {801--806},
pmid = {15965463},
publisher = {Nature Publishing Group},
title = {{Microstructure of a spatial map in the entorhinal cortex}},
url = {https://www.nature.com/articles/nature03721},
volume = {436},
year = {2005}
}

@article{Yamins2014,
author = {Yamins, Daniel L.K. and Hong, Ha and Cadieu, Charles F. and Solomon, Ethan A. and Seibert, Darren and DiCarlo, James J.},
doi = {10.1073/pnas.1403112111},
issn = {10916490},
journal = {Proceedings of the National Academy of Sciences of the United States of America},
month = {jun},
number = {23},
pages = {8619--8624},
pmid = {24812127},
publisher = {National Academy of Sciences},
title = {{Performance-optimized hierarchical models predict neural responses in higher visual cortex}},
url = {https://pubmed.ncbi.nlm.nih.gov/24812127/},
volume = {111},
year = {2014}
}

@article{Choudhary2020,
author = {Choudhary, Anirudh and Tong, Li and Zhu, Yuanda and Wang, May D.},
doi = {10.1055/s-0040-1702009},
issn = {23640502},
journal = {Yearbook of medical informatics},
number = {1},
pages = {129--138},
pmid = {32823306},
publisher = {Thieme Medical Publishers},
title = {{Advancing Medical Imaging Informatics by Deep Learning-Based Domain Adaptation}},
url = {/pmc/articles/PMC7442502/?report=abstract https://www.ncbi.nlm.nih.gov/pmc/articles/PMC7442502/},
volume = {29},
year = {2020}
}

@book{Piaget2006,
author = {Piaget, Jean},
booktitle = {The construction of reality in the child.},
doi = {10.1037/11168-000},
month = {oct},
publisher = {Basic Books},
title = {{The construction of reality in the child.}},
year = {2006}
}

@article{Hochreiter1997,
author = {Hochreiter, Sepp and Schmidhuber, J{\"{u}}rgen},
doi = {10.1162/neco.1997.9.8.1735},
issn = {08997667},
journal = {Neural Computation},
month = {nov},
number = {8},
pages = {1735--1780},
pmid = {9377276},
publisher = {MIT Press Journals},
title = {{Long Short-Term Memory}},
url = {https://www.mitpressjournals.org/doix/abs/10.1162/neco.1997.9.8.1735},
volume = {9},
year = {1997}
}

@article{Galama2019,
archivePrefix = {arXiv},
arxivId = {1804.05651},
author = {Galama, Ysbrand and Mensink, Thomas},
doi = {10.1016/j.cviu.2019.102803},
eprint = {1804.05651},
isbn = {2019.102803},
issn = {1090235X},
journal = {Computer Vision and Image Understanding},
title = {{IterGANs: Iterative GANs to learn and control 3D object transformation}},
url = {https://doi.org/10.1016/j.cviu.2019.102803},
volume = {189},
year = {2019}
}

@article{Osherson1990,
author = {Osherson, Daniel N. and Wilkie, Ormond and Smith, Edward E. and L{\'{o}}pez, Alejandro and Shafir, Eldar},
doi = {10.1037/0033-295X.97.2.185},
issn = {0033295X},
journal = {Psychological Review},
number = {2},
pages = {185--200},
publisher = {American Psychological Association Inc.},
title = {{Category-based induction}},
url = {/record/1990-18953-001},
volume = {97},
year = {1990}
}

@book{Marcus2001,
author = {Marcus, Gary F},
booktitle = {The Algebraic Mind},
doi = {10.7551/mitpress/1187.001.0001},
title = {{The Algebraic Mind}},
url = {http://cognet.mit.edu/book/algebraic-mind},
year = {2001}
}

@article{Chollet2019,
archivePrefix = {arXiv},
journal = {arXiv 1911.01547v2},
author = {Chollet, Fran{\c{c}}ois},
eprint = {1911.01547v2},
title = {{On the Measure of Intelligence}},
year = {2019}
}

@inproceedings{lake2018generalization,
  title={Generalization without systematicity: On the compositional skills of sequence-to-sequence recurrent networks},
  author={Lake, Brenden and Baroni, Marco},
  booktitle={International Conference on Machine Learning},
  pages={2873--2882},
  year={2018},
  organization={PMLR}
}

@article{summerfield2020structure,
  title={Structure learning and the posterior parietal cortex},
  author={Summerfield, Christopher and Luyckx, Fabrice and Sheahan, Hannah},
  journal={Progress in Neurobiology},
  volume={184},
  pages={101717},
  year={2020},
  publisher={Elsevier}
}

@inproceedings{santoro2018measuring,
  title={Measuring abstract reasoning in neural networks},
  author={Santoro, Adam and Hill, Felix and Barrett, David and Morcos, Ari and Lillicrap, Timothy},
  booktitle={International Conference on Machine Learning},
  pages={4477--4486},
  year={2018}
}

@article{lake2017building,
  title={Building machines that learn and think like people},
  author={Lake, Brenden M and Ullman, Tomer D and Tenenbaum, Joshua B and Gershman, Samuel J},
  journal={Behavioral and brain sciences},
  volume={40},
  year={2017},
  publisher={Cambridge University Press}
}

@article{attneave1956quantitative,
  title={The quantitative study of shape and pattern perception.},
  author={Attneave, Fred and Arnoult, Malcolm D},
  journal={Psychological bulletin},
  volume={53},
  number={6},
  pages={452},
  year={1956},
  publisher={American Psychological Association}
}

@article{xu2020neural,
  title={How Neural Networks Extrapolate: From Feedforward to Graph Neural Networks},
  author={Xu, Keyulu and Li, Jingling and Zhang, Mozhi and Du, Simon S and Kawarabayashi, Ken-ichi and Jegelka, Stefanie},
  journal={arXiv preprint arXiv:2009.11848},
  year={2020}
}

@article{hill2019emergent,
  title={Emergent systematic generalization in a situated agent},
  author={Hill, Felix and Lampinen, Andrew and Schneider, Rosalia and Clark, Stephen and Botvinick, Matthew and McClelland, James L and Santoro, Adam},
  journal={arXiv preprint arXiv:1910.00571},
  year={2019}
}

@article{cranmer2020lagrangian,
  title={Lagrangian neural networks},
  author={Cranmer, Miles and Greydanus, Sam and Hoyer, Stephan and Battaglia, Peter and Spergel, David and Ho, Shirley},
  journal={arXiv preprint arXiv:2003.04630},
  year={2020}
}

@inproceedings{greydanus2019hamiltonian,
  title={Hamiltonian neural networks},
  author={Greydanus, Samuel and Dzamba, Misko and Yosinski, Jason},
  booktitle={Advances in Neural Information Processing Systems},
  pages={15379--15389},
  year={2019}
}

@article{o2014learning,
  title={Learning through time in the thalamocortical loops},
  author={O'Reilly, Randall C and Wyatte, Dean and Rohrlich, John},
  journal={arXiv preprint arXiv:1407.3432},
  year={2014}
}

@article{mehrabi2019survey,
  title={A survey on bias and fairness in machine learning},
  author={Mehrabi, Ninareh and Morstatter, Fred and Saxena, Nripsuta and Lerman, Kristina and Galstyan, Aram},
  journal={arXiv preprint arXiv:1908.09635},
  year={2019}
}

@article{lecun1995convolutional,
  title={Convolutional networks for images, speech, and time series},
  author={LeCun, Yann and Bengio, Yoshua and others},
  journal={The handbook of brain theory and neural networks},
  volume={3361},
  number={10},
  pages={1995},
  year={1995}
}

@article{rao1999predictive,
  title={Predictive coding in the visual cortex: a functional interpretation of some extra-classical receptive-field effects},
  author={Rao, Rajesh PN and Ballard, Dana H},
  journal={Nature neuroscience},
  volume={2},
  number={1},
  pages={79--87},
  year={1999},
  publisher={Nature Publishing Group}
}

@article{wolpert1995internal,
  title={An internal model for sensorimotor integration},
  author={Wolpert, Daniel M and Ghahramani, Zoubin and Jordan, Michael I},
  journal={Science},
  volume={269},
  number={5232},
  pages={1880--1882},
  year={1995},
  publisher={American Association for the Advancement of Science}
}

\newpage

\appendix
\counterwithin{figure}{section}

\section{Supplementary Figures}

\begin{figure}[htp]
  \includegraphics{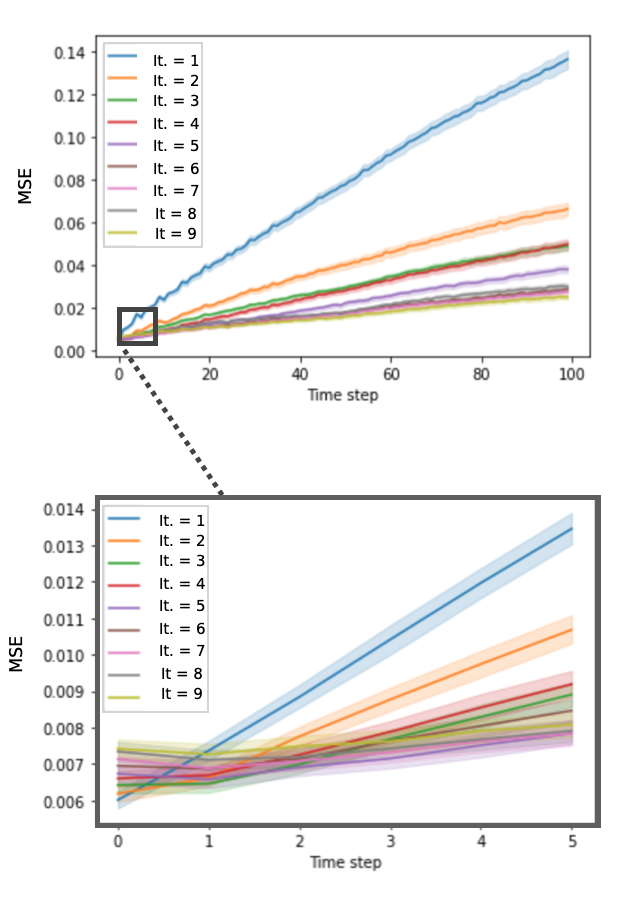}
  \caption{Top: Mean squared error (MSE) on I.I.D. test set at each time step for 9 trained rotation networks. The label ‘It. = N’ indicates that during training, the number of iterations of the newtork on a given batch was sampled uniformly between 1 and N. Bottom: the same plot blown up to visualize the first 6 time-steps. Of note, networks trained with higher iterations actually have a worse MSE at the first time step, but achieve a much better MSE in the long run. Confidence intervals represent standard error between 3 identically trained networks.}
  \label{figa1}
\end{figure}

\begin{figure}[htp]
  \centering
    \resizebox{\textwidth}{!}{ }
  \includegraphics{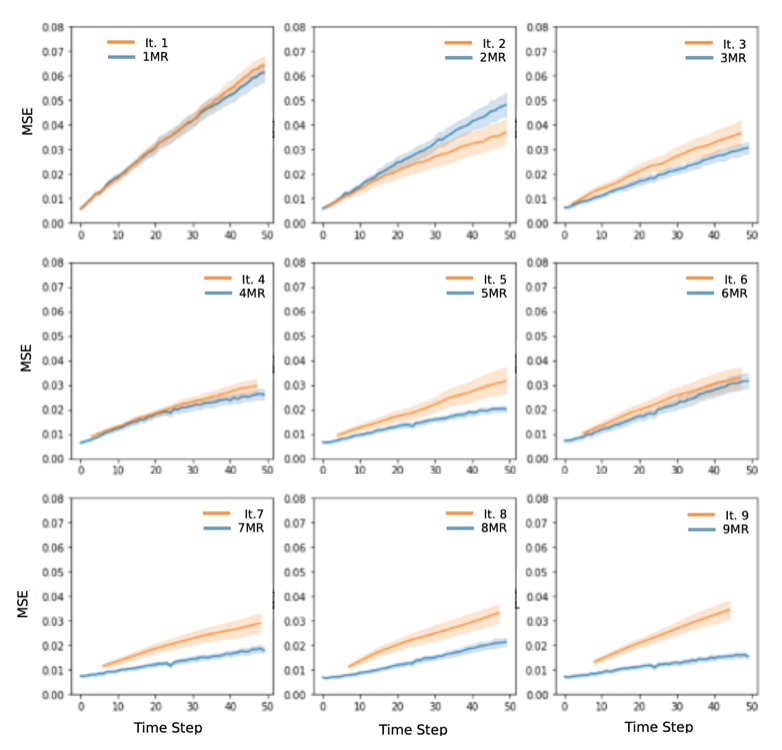}
  \caption{Comparison of iteravely trained rotation networks (It. 1 through 9) with networks that were trained to rotate a specified angular distance in a single forward pass. For example, if the network It. 5 experienced between 1 and 5 iterative passes during training, then the comparison network was trained to rotate an object by 5 minimal rotations (5MR or $\frac{5\pi}{25}$ radians) in a single pass. For higher values of iterations/MR, the iteravely trained networks have improved performance over time, as indicated by mean squared error loss (MSE), even though they are rotating objects using many more steps (1 minimal rotation at a time). Confidence intervals represent standard error between 3 identically trained networks.}
\end{figure}

\end{document}